\documentclass[sigconf]{acmart}
\usepackage{multirow}
\usepackage{makecell}
\usepackage{graphicx}
\usepackage{caption}
\usepackage{subfigure}
\usepackage{adjustbox}
\AtBeginDocument{%
  \providecommand\BibTeX{{%
    \normalfont B\kern-0.5em{\scshape i\kern-0.25em b}\kern-0.8em\TeX}}}

\copyrightyear{2025}
\acmYear{2025}

\setcopyright{acmlicensed}\acmConference[CIKM '25]{Proceedings of the 34th ACM International Conference on Information and Knowledge Management (CIKM ’25)}{November 10-14, 2025}{Seoul, Republic of Korea}
\acmBooktitle{Proceedings of the 34th ACM International Conference on Information and Knowledge Management (CIKM ’25), November 10-14, 2025, Seoul, Republic of Korea}
\acmDOI{10.1145/3746252.3761643}
\acmISBN{979-8-4007-2040-6/2025/11.}

\begin{document}

\title{FinS-Pilot: A Benchmark for Online Financial RAG System}



\author{Feng Wang}
\affiliation{%
  \institution{Gaoling School of Artificial Intelligence, Renmin University of China}
  \institution{Beijing Key Laboratory of Research on Large Models and Intelligent Governance}
  \institution{Engineering Research Center of Next-Generation Intelligent Search and Recommendation, MOE}
  \city{Beijing}
  \country{China}
}
\email{rucwangfeng@gmail.com}

\author{Yiding Sun}
\affiliation{%
  \institution{Gaoling School of Artificial Intelligence, Renmin University of China}
  \institution{Beijing Key Laboratory of Research on Large Models and Intelligent Governance}
  \institution{Engineering Research Center of Next-Generation Intelligent Search and Recommendation, MOE}
  \institution{Tencent Inc.}
  \city{Beijing}
  \country{China}
}
\email{emanual20.sun@gmail.com}

\author{Jiaxin Mao\textsuperscript{*}}
\affiliation{%
  \institution{Gaoling School of Artificial Intelligence, Renmin University of China}
  \institution{Beijing Key Laboratory of Research on Large Models and Intelligent Governance}
  \institution{Engineering Research Center of Next-Generation Intelligent Search and Recommendation, MOE}
  \city{Beijing}
  \country{China}
}
\email{maojiaxin@gmail.com}

\author{Wei Xue}
\affiliation{%
  \institution{Zhicepilot}
  \city{Beijing}
  \country{China}
}
\email{xuewei@zhicepilot.com}

\author{Danqing Xu}
\affiliation{%
  \institution{Zhicepilot}
  \city{Beijing}
  \country{China}
}
\email{xudanqing@zhicepilot.com}


\renewcommand{\shortauthors}{Feng Wang, Yiding Sun, Jiaxin Mao, Wei Xue, \& Danqing Xu}

\begin{abstract}


Large language models (LLMs) have demonstrated remarkable capabilities across various professional domains, with their performance typically evaluated through standardized benchmarks. 
In the financial field, the stringent demands for professional accuracy and real-time data processing often necessitate the use of retrieval-augmented generation (RAG) techniques.
However, the development of financial RAG benchmarks has been constrained by data confidentiality issues and the lack of dynamic data integration. 
To address this issue, we introduce FinS-Pilot, a novel benchmark for evaluating RAG systems in online financial applications. Constructed from real-world financial assistant interactions, our benchmark incorporates both real-time API data and text data, organized through an intent classification framework covering critical financial domains. 
The benchmark enables comprehensive evaluation of financial assistants' capabilities in handling both static knowledge and time-sensitive market information.
Through systematic experiments with multiple Chinese leading LLMs, we demonstrate FinS-Pilot's effectiveness in identifying models suitable for financial applications while addressing the current gap in specialized evaluation tools for the financial domain. 
Our work contributes both a practical evaluation framework and a curated dataset to advance research in financial NLP systems. The code and dataset are accessible on GitHub\footnote{https://github.com/PhealenWang/financial\_rag\_benchmark}.

\end{abstract}

\ccsdesc[500]{Computing methodologies~Natural language processing}

\keywords{Retrieval-Augmented Generation; Financial; Benchmark}

\maketitle

\let\thefootnote\relax\footnotetext{* Corresponding author.}

\section{Introduction}

\begin{figure*}[ht]
  \centering
  \includegraphics[width=0.82\linewidth]{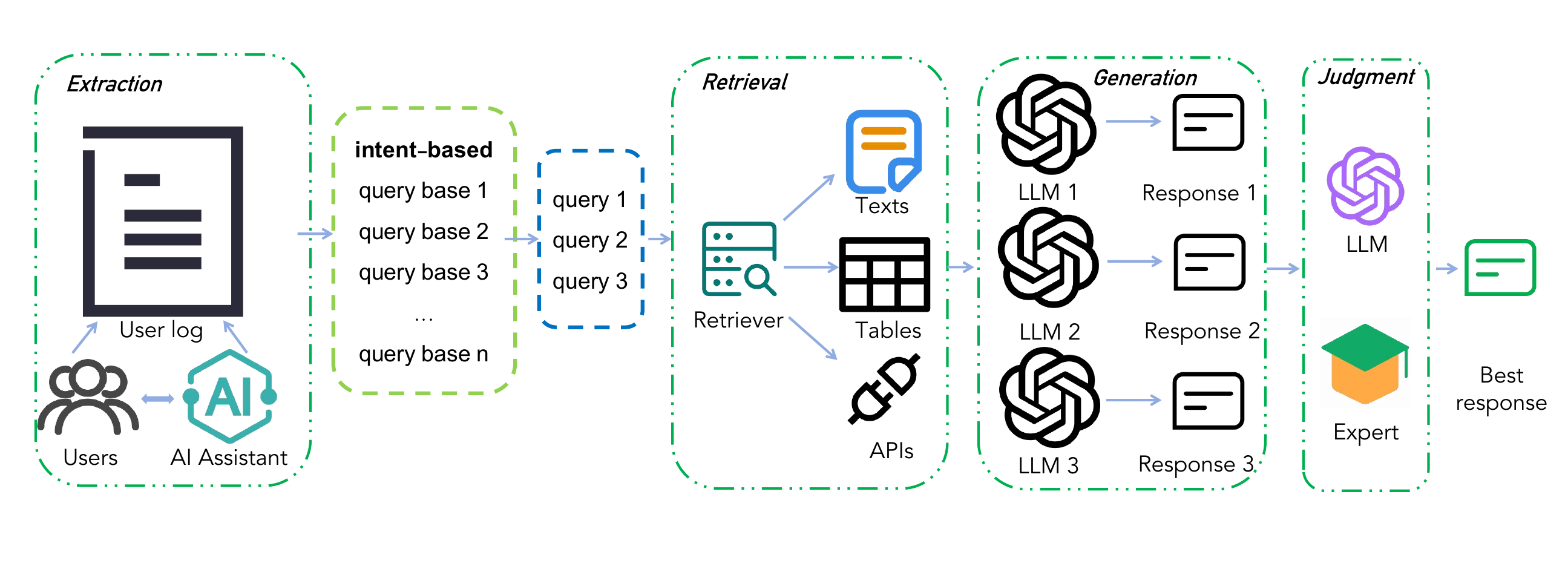}
  \setlength{\abovecaptionskip}{-0.6cm}
  \setlength{\belowcaptionskip}{-0.8cm}
  \caption{Overview of constructing process.}
  \label{fig:financial-workflow}
  \vspace{0.4cm}
\end{figure*}


LLMs have profoundly transformed social production and people's working patterns. Their remarkable information comprehension and generation capabilities endow LLMs with extensive application value. In both academia and industry, LLMs have achieved significant performance improvements in multiple tasks within the fields of NLP and IR. In the field of IR, LLMs and traditional information retrieval technologies have started to influence and collaborate with each other, resulting the retrieval-augmented generation technology~(RAG)~\cite{mei2025survey}. RAG provides LLMs with highly professional and real-time external knowledge through information retrieval, reducing hallucinations during the generation process of LLMs~\cite{RAG-hallucination-benchmark-AAAI2024,chen2025improving,chen2025mao,guo2025deepseek}. RAG is widely applied in fields with strong professionalism and real-time requirements, and the financial field is one of them.


Since the datasets in the financial field may involve business secrets and have strong privacy and high acquisition difficulty problems, the current benchmarks for RAG in the financial field are still in short supply. Therefore, constructing an open-source benchmark for financial RAG is important for evaluation.
Currently, the evaluation datasets for financial tasks include FinanceBench~\cite{islam2023financebench}, FiQA~\cite{maia201818}, FinQA~\cite{chen2021finqa}, etc., and they still have many problems.


Firstly, these datasets are constructed from financial report, with queries focused on financial question-answering tasks based on static reports. 
However, the low update frequency of financial reports leads to a lack of real-time information.
Secondly, they do not utilize dynamic data interfaces and cannot obtain real-time market information. 
Moreover, the queries are not from the real application scenarios, resulting the gap between their evaluation results and the performance of models in the online environment.

To address these issues, we construct a financial RAG benchmark. 
Besides text data, we also provide APIs for accessing real-time market data.
The contributions of our financial RAG benchmark can be summarized as the following:
\begin{itemize}
\setlength{\itemsep}{0pt}
\setlength{\parsep}{0pt}
\setlength{\parskip}{0pt}
\item \textbf{Online user-driven data collection}. The queries in the benchmark are obtained from the user logs of an online financial assistant. Through rigorous desensitization and context-preserving adaptation processes, we maintain 316 queries spanning investment advice, regulatory compliance, and transactional inquiries—representing real-world usage patterns absent in synthetic benchmarks.

\item \textbf{Integration of dynamic data sources}. Beyond conventional static text corpora, we incorporate real-time API responses from market data and financial calculation services. 
This hybrid approach enables more realistic evaluation of RAG systems' capacity to handle both static knowledge and time-sensitive numerical data, addressing a critical gap in existing evaluation methodologies.

\item \textbf{Workflow-aware intent taxonomy}. Mirroring the operational classification system used by professional financial assistants, our dataset organizes questions into 62 functionally distinct intent categories. This structure enables a fine-grained analysis of LLM performance across different scenarios, with evaluation metrics specifically designed to inform pipeline optimization decisions in production environments.

\end{itemize} 


\section{Benchmark Construction}
\label{sec:construction}


The benchmark construction pipeline is illustrated in Figure~\ref{fig:financial-workflow}, comprising four steps.
The first step is \textbf{Extraction}.
Online financial assistant interactions generate logs containing authentic queries and corresponding responses. Leveraging workflow-specific patterns, we perform two-level intent classification of queries aligned with actual business processes. Through systematic clustering of queries within each second-level category, we establish query bases, where each intent maps to one or more query classes, which subsequently group semantically similar user questions.
The second step is \textbf{Retrieval}. 
To address the domain-specific knowledge requirements of financial queries, we implement a dual-source retrieval system:
(i) Real-time dynamic data via the Tushare API for time-sensitive financial metrics, and
(ii) Structured text databases retrieved through embedding-based methods.
This hybrid approach ensures comprehensive reference documentation covering both static knowledge and live market data.
The third step is \textbf{Generation}.
Using the collected queries and retrieved documents, we prompt multiple Chinese LLMs to produce candidate answers. These outputs serve as preliminary responses for subsequent evaluation.
The final step is \textbf{Judgment}. 
While LLMs generate candidate answers, we employ a two-tier verification process:
(i) An LLM-based discriminator identifies the most semantically aligned candidate with expected responses, followed by
(ii) Manual review by cross-disciplinary experts (finance + AI) to finalize gold-standard answers.

\subsection{Extraction}

\noindent\textbf{Dataset source and preprocessing.}\quad
The user interaction logs from our financial assistant system contain three key elements: user queries, system responses, and precise timestamps. These logs undergo careful preprocessing to remove sensitive information.

\noindent\textbf{Workflow-based classification.}\quad
Following the operational practices of financial assistants, we develop a hierarchical classification system with 9 first-level categories (e.g., macroeconomic analysis, market analysis) and 62 second-level categories (e.g., macroeconomic analysis, market sentiment). Each second-level category corresponds to specific business workflows, enabling precise routing of different query types to appropriate processing pipelines.

\noindent\textbf{Query base abstraction.}\quad
We observe most user queries follow repetitive patterns with variations (e.g. time, entity names). For each second-level category, we create query bases that capture these patterns. For example, the query base \texttt{[Company][Time][Valuation Status]} abstracts various forms of stock valuation questions. These templates systematically incorporate temporal dimensions to reflect financial information's time-sensitive nature.

\noindent\textbf{Dataset composition.}\quad
We retain 316 queries and divide them into numerical queries (104 in total, with dynamic interface data as relevant information, specifically concentrated under 4 secondary intents) and content-based queries (212 in total, with text data as relevant information) according to the types of answers and relevant information.

\subsection{Retrieval}


The retrieval process is a critical component of our framework, where each user query requires reference materials. We implement a dual-channel retrieval system that combines: (1) real-time dynamic data interfaces for numerical queries, and (2) comprehensive text corpora for content-based queries. This approach ensures optimal information timeliness for time-sensitive financial data while maintaining broad knowledge coverage for conceptual questions.

\noindent\textbf{Dynamic Interface Data Implementation.}\quad
Text data is difficult to provide real-time market information (e.g., stock prices, corporate financial indicators). Therefore, we employ the Tushare Pro API as our primary data source. Tushare provides professionally curated financial datasets spanning multiple asset classes including equities, funds, futures, and cryptocurrencies.
Our implementation involves (i)~automated API calls triggered by query classification, (ii)~structured response parsing into standardized JSON formats, and (iii)~quality validation through checksum verification.

\noindent\textbf{Text Data Acquisition and Processing.}\quad
The text data comes from news, research reports, and the Internet. To ensure timeliness, it will be updated regularly. Relevant texts are retrieved from the database through relevance retrieval. Since the knowledge in the text database is limited, we introduce the Bing search interface to provide more comprehensive text data for user queries. 

\subsection{Generation}


The generation phase is the core of our RAG pipeline. Leveraging the instruction-following capabilities of the LLMs, we design task-specific prompts that incorporate both the retrieved documents and structured instructions. Our prompt engineering follows three principles: (1) explicit task specification, (2) clear input-output formatting, and (3) contextual grounding in financial domain knowledge.

\textbf{Model Selection}. 
We select five popular Chinese LLMs, including DeepSeek-v3~\cite{liu2024deepseek}, DeepSeek-R1~\cite{guo2025deepseek}, Doubao-1.5-pro, Moonshot-v1, and Baichuan4, to perform the generation task. We use them to generate answers for each query respectively, forming a candidate set of standard answers.

\textbf{Generation Processing}. 
For numerical queries, we provide dynamic interface data to the LLMs in pandas' Dataframe format, and attach explanatory texts about the meanings of data elements. For content-based queries, we incorporate user queries and reference documents into the prompts, enabling the LLMs to generate corresponding answers based on them. 

\subsection{Judgment}


The process of selecting optimal answers from candidate texts requires distinct approaches for numerical and content-based queries, reflecting their fundamental differences in information characteristics and evaluation criteria. 
For numerical queries where ground truth values exist in the retrieved dynamic interface data, we implement a rigorous manual verification protocol. we directly invite professionals with rich experience in data processing and proficient in data processingy to extract values from the structured API responses, with final values determined through consensus when discrepancies occur. This manual extraction process ensures 100\% accuracy for time-sensitive numerical indicators that might be misinterpreted by automated systems.

For content-based queries, we first use a large language model (Doubao-1.5-pro) to judge the relevance between all candidate answers and user queries, and selects the most relevant answer as the only candidate for the standard answer. Subsequently, we first remove the content in the answer that is irrelevant to the user query, and then invites experts in the fields of artificial intelligence and finance to review, correct the wrong information therein, and finally obtain the standard answer for each query. 

The final validation involves cross-disciplinary review by two AI researchers and two financial domain experts independently. They evaluate each candidate answer against four quality dimensions: factual accuracy, financial terminology appropriateness, logical flow, and compliance with financial reporting standards. Disagreements are resolved through discussion until consensus is reached.

\subsection{Overview}

We provide a benchmark including 104 numerical queries and 212 content-based queries. 
Each query comes with a manually verified groundtruth answer.
We also provide a corpus containing relevant information for content-based queries, some APIs to retrieve stock price data, macroeconomic indicators, fund data for answering numerical queries, and automated evaluation code.
More usage details, such as query cases and the taxonomy of intention classification, are available on our github repository.

\section{Experiments}
\label{sec:experiments}

In this section, we present our experiment settings and results for numerical queries and content-based queries.

\subsection{Settings}
\label{exp:settings}

\noindent\textbf{Generators.}\quad
We use popular Chinese LLMs, including DeepSeek-v3~\cite{liu2024deepseek}, DeepSeek-R1~\cite{guo2025deepseek}, Doubao-1.5-pro, Moonshot-v1, Baichuan4, and the closed source model Xiaofa-1.0 from Zhicepilot to generate answers for all the queries. 

\noindent\textbf{Retrievers.}\quad
We configure three distinct retriever categories including (1)~\textbf{Base}, an embedding-based dense retriever leveraging neural representations; (2)~\textbf{Bing}, which incorporates external web search capabilities through Microsoft's API; (3)~\textbf{Bert}~\cite{devlin2018bert} without fine-tuning; and (4)~\textbf{Close} as a non-retrieval baseline that processes queries without any information retrieval component.

\noindent\textbf{Metrics.}\quad
We have designed evaluation metrics for two query types respectively. For numerical queries, we use accuracy as the metric. For content-based queries, we have provided similarity-based metrics and model-based metrics. The similarity-based metrics include ROUGE-L (ROU.)~\cite{lin2004rouge}, BLEU~\cite{papineni2002bleu} and cosine similarity (COS.). The model-based metrics include hallucination (HAL.), completeness (COM.), and relevance (REL.).  
Hallucination assesses whether the response contains information conflicting with references. Completeness measures the extent to which the response covers all subtopics addressed in the ground truth. Relevance evaluates the alignment between the response and the query intent.

\subsection{Results}

\begin{figure}[t]
    \centering
    \includegraphics[width=0.45\textwidth]{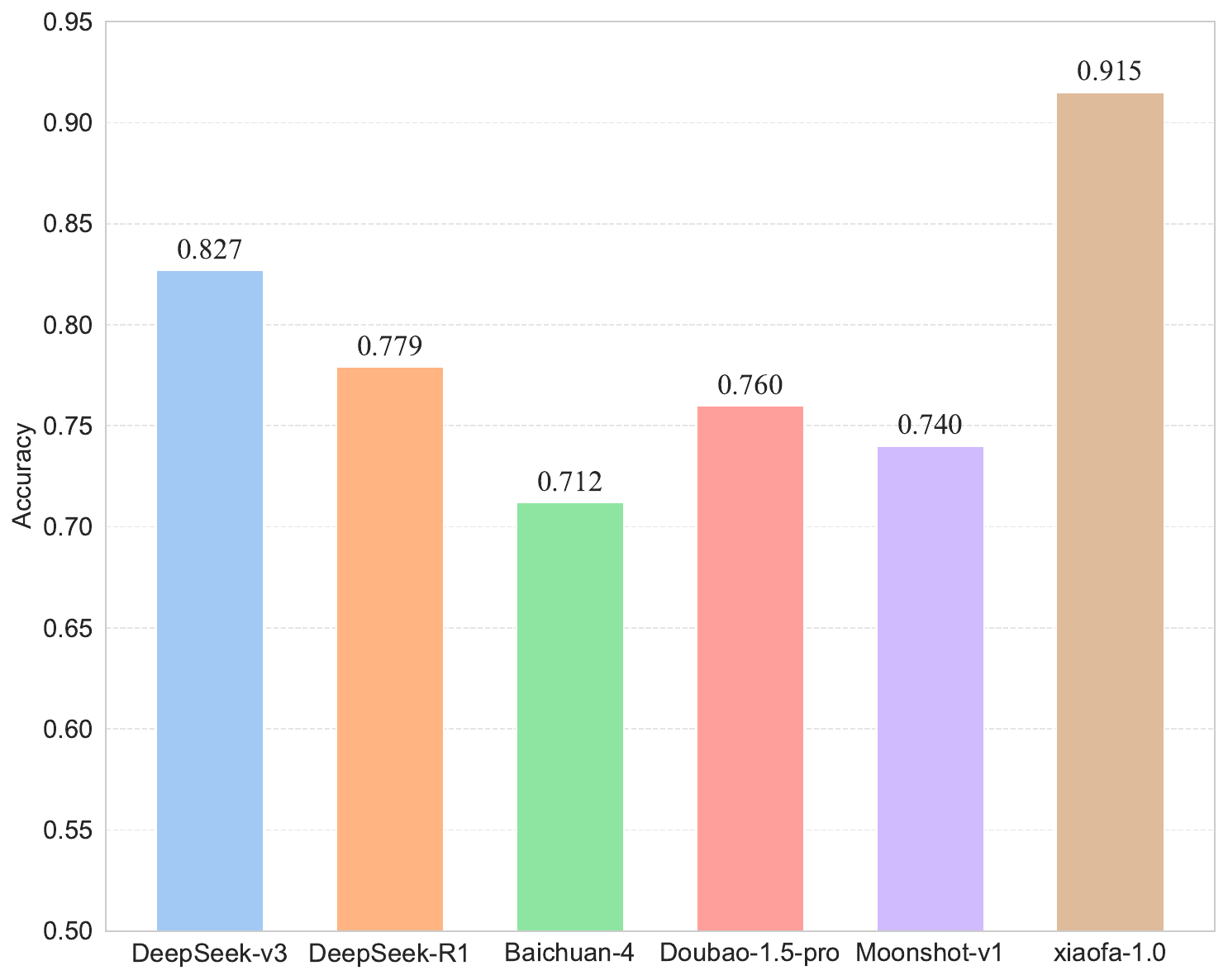}
    \caption{Evaluation of numerical queries}
    \label{fig:value_query_evaluation}
    \vspace{-0.7cm}
\end{figure}

\noindent\textbf{Numerical queries.}\quad
We process 104 numerical queries DeepSeek-v3 without any reference data, yields zero accuracy, which reveals the necessary requirement for real-time data integration in financial applications. Other results are presented in Figure~\ref{fig:value_query_evaluation}.

Xiaofa-1.0 emerges as the top-performing model, achieving 91.5\% accuracy. This performance suggests superior capabilities in both structured data interpretation and numerical reasoning compared to other LLMs. The remaining LLMs demonstrate comparable performance within the 71\%-83\% range, indicating adequate but not exceptional proficiency in financial data processing.
Error analysis reveals two primary failure modes: unit conversion mistakes and temporal reference misinterpretations.

\noindent\textbf{Content-based queries.}\quad
We process 212 content-based queries through the generation pipeline described in Section~\ref{exp:settings}. The comprehensive evaluation results, measured across all specified metrics, are presented in Table~\ref{tab:content-evaluation}.
To ensure consistent interpretation across evaluation dimensions, we apply maximum normalization to both the Completeness (COM.) and Relevance (REL.) metrics. This preprocessing step linearly rescales the raw scores by a factor of 1/4, transforming all metric values to a standardized [0,1] interval.

Overall, Xiaofa-1.0 achieves best results in most metrics under Base and Bing setup, which indicates that Xiaofa-1.0 can effectively complete content-based generation tasks in scenarios with a limited database and the introduction of external document data.

The systematic comparison of four retrieval paradigms yields several insights: 
(i) LLMs consistently perform better under Bing setting compared to Base. This gap reveals the value of incorporating external reference materials for enhanced generation quality.
(ii) The results under Close and Bert settings perform poorly in all metrics, empirically validating the necessity of retrieval augmentation for financial assistant applications.
These findings demonstrate while all models benefit from explicit retrieval augmentation, their strengths vary significantly based on both architectural differences and knowledge source availability. The results particularly highlight Xiaofa-1.0's robustness across constrained environments and superior external data integration capabilities.

\begin{table}[t]
    \centering
    \small
    \caption{Evaluation of content-based queries}
    \setlength{\tabcolsep}{1.5mm}
    \begin{adjustbox}{width=0.47\textwidth}
    \begin{tabular}{cccccccc}
        \toprule
        \textbf{Retriever} & \textbf{Generator} & \textbf{ROU.}$\uparrow$ & \textbf{BLEU}$\uparrow$ & \textbf{COS.}$\uparrow$ & \textbf{HAL.}$\downarrow$ & \textbf{COM.}$\uparrow$ & \textbf{REL.}$\uparrow$ \\
        \midrule
        \multirow{6}{*}{\makecell[l]{Base}}
        & DeepSeek-v3 & .0612 & .0564 & \textbf{.3159} & .5423 & .2152 & .9780 \\
        & DeepSeek-R1 &  .0124 & .0321 & .2359 & .5771 & .2077 & .9588 \\
        & Baichuan-4 & .0361 & .0358 & .2847 & .5970 & .1841 & .9007 \\
        & Doubao-1.5-pro & .0387 & .0506 & .2508 & .4726 & .1791 & .8634 \\
        & Moonshot-v1 & .0504 & .0537 & .2526 & .4975 & .1642 & .9531 \\
        & Xiaofa-1.0 (Ours) & \textbf{.0696} & \textbf{.0845} & .3070 & \textbf{.2008} & \textbf{.2282} & \textbf{.9814} \\
        \midrule
        \multirow{6}{*}{\makecell[l]{Bing}}
        & DeepSeek-v3 & .1676 & .0647 & .5480 & .0746 & .5323 & .9774 \\
        & DeepSeek-R1 &  .2098 & .0533 & .5056 & .0697 & .5522 & .9536 \\
        & Baichuan-4 & .2094 & .0458 & .5430 & .1194 & .5373 & .9714 \\
        & Doubao-1.5-pro & \textbf{.3469} & .0467 & \textbf{.6080} & .0547 & .6132 & .9333 \\
        & Moonshot-v1 & .2582 & .0960 & .5515 & .1045 & .5758 & .9684 \\
        & Xiaofa-1.0 (Ours) & .2043 & \textbf{.0979} & .5857 & \textbf{.0503} & \textbf{.6375} & \textbf{.9792} \\
        \midrule
        Bert & DeepSeek-v3 & .0022 & .1010 & .1570 & .0334 & .0177 & .1998 \\
        \midrule
        Close & DeepSeek-v3 & .0960 & .0714 & .3278 & .7363 & .1853 & .6500 \\
        \bottomrule
    \end{tabular}
    \end{adjustbox}
    \label{tab:content-evaluation}
    \vspace{-0.6cm}
\end{table}

\section{Related Works}
\label{sec:related}

\noindent\textbf{General-Purpose Benchmarking.}\quad
Current LLMs are typically evaluated across three core capabilities: language modeling, knowledge utilization, and natural language understanding~\cite{zhu2024yulan,sun2024integrated}. The LAMBDA dataset~\cite{lambada} serves as the standard for assessing contextual language modeling proficiency. For knowledge-intensive tasks, MMLU~\cite{hendrycks2020measuring} provides a comprehensive benchmark covering 57 subjects. The GLUE benchmark~\cite{wang2018glue} further enables multi-dimensional evaluation of linguistic understanding through diverse tasks including sentiment analysis and textual entailment.

\noindent\textbf{Domain-Specific Evaluation in Finance.}\quad 
The rapid adoption of LLMs in specialized domains has necessitated the development of vertical-specific evaluation frameworks. Financial applications present unique challenges: (1) proprietary data restrictions limit training corpus availability, (2) real-time market data integration is operationally critical~\cite{livebench, latesteval}, and (3) regulatory compliance requirements demand exceptional factual precision.

Existing financial benchmarks exhibit three key limitations. FinanceBench~\cite{islam2023financebench} and FinQA~\cite{chen2021finqa} focus narrowly on static financial statement analysis, without evaluation of real-time queries. While FiQA~\cite{maia201818} incorporates temporal aspects, its synthetic queries fail to capture authentic user intent distributions observed in production systems. Most critically, none of these benchmarks assess RAG capabilities - the predominant architectural pattern for financial LLM deployments due to its ability to combine parametric knowledge with dynamic data feeds.



\section*{ACKNOWLEDGEMENTS}

This research was supported by 
the Natural Science Foundation of China (61902209, 62377044), 
Intelligent Social Governance Platform, 
Major Innovation \& Planning Interdisciplinary Platform for the ``Double-First Class" Initiative, 
Beijing Nova Program, 
Renmin University of China, 
the Fundamental Research Funds for the Central Universities, and the Research Funds of Renmin University of China (22XNKJ15). 
Any opinions, findings, conclusions, or recommendations expressed in this material are those of the authors and do not necessarily reflect those of the sponsors.

\section*{GenAI Usage Disclosure}
In this work, GenAI was only utilized to help to generate candidate answers and select best answer in Section~\ref{sec:construction}. We manually refined the selected answers carefully.

\bibliographystyle{ACM-Reference-Format}
\bibliography{reference}


\begin{thebibliography}{19}


\ifx \showCODEN    \undefined \def \showCODEN     #1{\unskip}     \fi
\ifx \showDOI      \undefined \def \showDOI       #1{#1}\fi
\ifx \showISBNx    \undefined \def \showISBNx     #1{\unskip}     \fi
\ifx \showISBNxiii \undefined \def \showISBNxiii  #1{\unskip}     \fi
\ifx \showISSN     \undefined \def \showISSN      #1{\unskip}     \fi
\ifx \showLCCN     \undefined \def \showLCCN      #1{\unskip}     \fi
\ifx \shownote     \undefined \def \shownote      #1{#1}          \fi
\ifx \showarticletitle \undefined \def \showarticletitle #1{#1}   \fi
\ifx \showURL      \undefined \def \showURL       {\relax}        \fi
\providecommand\bibfield[2]{#2}
\providecommand\bibinfo[2]{#2}
\providecommand\natexlab[1]{#1}
\providecommand\showeprint[2][]{arXiv:#2}

\bibitem[Chen et~al\mbox{.}(2023)]%
        {RAG-hallucination-benchmark-AAAI2024}
\bibfield{author}{\bibinfo{person}{Jiawei Chen}, \bibinfo{person}{Hongyu Lin}, \bibinfo{person}{Xianpei Han}, {and} \bibinfo{person}{Le Sun}.} \bibinfo{year}{2023}\natexlab{}.
\newblock \showarticletitle{Benchmarking Large Language Models in Retrieval-Augmented Generation}.
\newblock \bibinfo{journal}{\emph{CoRR}}  \bibinfo{volume}{abs/2309.01431} (\bibinfo{year}{2023}).
\newblock
\urldef\tempurl%
\url{https://doi.org/10.48550/ARXIV.2309.01431}
\showDOI{\tempurl}
\showeprint[arXiv]{2309.01431}


\bibitem[Chen et~al\mbox{.}(2025a)]%
        {chen2025improving}
\bibfield{author}{\bibinfo{person}{Yiqun Chen}, \bibinfo{person}{Lingyong Yan}, \bibinfo{person}{Weiwei Sun}, \bibinfo{person}{Xinyu Ma}, \bibinfo{person}{Yi Zhang}, \bibinfo{person}{Shuaiqiang Wang}, \bibinfo{person}{Dawei Yin}, \bibinfo{person}{Yiming Yang}, {and} \bibinfo{person}{Jiaxin Mao}.} \bibinfo{year}{2025}\natexlab{a}.
\newblock \showarticletitle{Improving retrieval-augmented generation through multi-agent reinforcement learning}.
\newblock \bibinfo{journal}{\emph{arXiv preprint arXiv:2501.15228}} (\bibinfo{year}{2025}).
\newblock


\bibitem[Chen et~al\mbox{.}(2025b)]%
        {chen2025mao}
\bibfield{author}{\bibinfo{person}{Yiqun Chen}, \bibinfo{person}{Erhan Zhang}, \bibinfo{person}{Lingyong Yan}, \bibinfo{person}{Shuaiqiang Wang}, \bibinfo{person}{Jizhou Huang}, \bibinfo{person}{Dawei Yin}, {and} \bibinfo{person}{Jiaxin Mao}.} \bibinfo{year}{2025}\natexlab{b}.
\newblock \showarticletitle{MAO-ARAG: Multi-Agent Orchestration for Adaptive Retrieval-Augmented Generation}.
\newblock \bibinfo{journal}{\emph{arXiv preprint arXiv:2508.01005}} (\bibinfo{year}{2025}).
\newblock


\bibitem[Chen et~al\mbox{.}(2021)]%
        {chen2021finqa}
\bibfield{author}{\bibinfo{person}{Zhiyu Chen}, \bibinfo{person}{Wenhu Chen}, \bibinfo{person}{Charese Smiley}, \bibinfo{person}{Sameena Shah}, \bibinfo{person}{Iana Borova}, \bibinfo{person}{Dylan Langdon}, \bibinfo{person}{Reema Moussa}, \bibinfo{person}{Matt Beane}, \bibinfo{person}{Ting-Hao Huang}, \bibinfo{person}{Bryan Routledge}, {et~al\mbox{.}}} \bibinfo{year}{2021}\natexlab{}.
\newblock \showarticletitle{Finqa: A dataset of numerical reasoning over financial data}.
\newblock \bibinfo{journal}{\emph{arXiv preprint arXiv:2109.00122}} (\bibinfo{year}{2021}).
\newblock


\bibitem[Devlin(2018)]%
        {devlin2018bert}
\bibfield{author}{\bibinfo{person}{Jacob Devlin}.} \bibinfo{year}{2018}\natexlab{}.
\newblock \showarticletitle{Bert: Pre-training of deep bidirectional transformers for language understanding}.
\newblock \bibinfo{journal}{\emph{arXiv preprint arXiv:1810.04805}} (\bibinfo{year}{2018}).
\newblock


\bibitem[Guo et~al\mbox{.}(2025)]%
        {guo2025deepseek}
\bibfield{author}{\bibinfo{person}{Daya Guo}, \bibinfo{person}{Dejian Yang}, \bibinfo{person}{Haowei Zhang}, \bibinfo{person}{Junxiao Song}, \bibinfo{person}{Ruoyu Zhang}, \bibinfo{person}{Runxin Xu}, \bibinfo{person}{Qihao Zhu}, \bibinfo{person}{Shirong Ma}, \bibinfo{person}{Peiyi Wang}, \bibinfo{person}{Xiao Bi}, {et~al\mbox{.}}} \bibinfo{year}{2025}\natexlab{}.
\newblock \showarticletitle{Deepseek-r1: Incentivizing reasoning capability in llms via reinforcement learning}.
\newblock \bibinfo{journal}{\emph{arXiv preprint arXiv:2501.12948}} (\bibinfo{year}{2025}).
\newblock


\bibitem[Hendrycks et~al\mbox{.}(2020)]%
        {hendrycks2020measuring}
\bibfield{author}{\bibinfo{person}{Dan Hendrycks}, \bibinfo{person}{Collin Burns}, \bibinfo{person}{Steven Basart}, \bibinfo{person}{Andy Zou}, \bibinfo{person}{Mantas Mazeika}, \bibinfo{person}{Dawn Song}, {and} \bibinfo{person}{Jacob Steinhardt}.} \bibinfo{year}{2020}\natexlab{}.
\newblock \showarticletitle{Measuring massive multitask language understanding}.
\newblock \bibinfo{journal}{\emph{arXiv preprint arXiv:2009.03300}} (\bibinfo{year}{2020}).
\newblock


\bibitem[Islam et~al\mbox{.}(2023)]%
        {islam2023financebench}
\bibfield{author}{\bibinfo{person}{Pranab Islam}, \bibinfo{person}{Anand Kannappan}, \bibinfo{person}{Douwe Kiela}, \bibinfo{person}{Rebecca Qian}, \bibinfo{person}{Nino Scherrer}, {and} \bibinfo{person}{Bertie Vidgen}.} \bibinfo{year}{2023}\natexlab{}.
\newblock \showarticletitle{Financebench: A new benchmark for financial question answering}.
\newblock \bibinfo{journal}{\emph{arXiv preprint arXiv:2311.11944}} (\bibinfo{year}{2023}).
\newblock


\bibitem[Li et~al\mbox{.}(2024)]%
        {latesteval}
\bibfield{author}{\bibinfo{person}{Yucheng Li}, \bibinfo{person}{Frank Guerin}, {and} \bibinfo{person}{Chenghua Lin}.} \bibinfo{year}{2024}\natexlab{}.
\newblock \showarticletitle{LatestEval: addressing data contamination in language model evaluation through dynamic and time-sensitive test construction}. In \bibinfo{booktitle}{\emph{Proceedings of the Thirty-Eighth AAAI Conference on Artificial Intelligence and Thirty-Sixth Conference on Innovative Applications of Artificial Intelligence and Fourteenth Symposium on Educational Advances in Artificial Intelligence}} \emph{(\bibinfo{series}{AAAI'24/IAAI'24/EAAI'24})}. \bibinfo{publisher}{AAAI Press}, Article \bibinfo{articleno}{2074}, \bibinfo{numpages}{8}~pages.
\newblock
\showISBNx{978-1-57735-887-9}
\urldef\tempurl%
\url{https://doi.org/10.1609/aaai.v38i17.29822}
\showDOI{\tempurl}


\bibitem[Lin(2004)]%
        {lin2004rouge}
\bibfield{author}{\bibinfo{person}{Chin-Yew Lin}.} \bibinfo{year}{2004}\natexlab{}.
\newblock \showarticletitle{Rouge: A package for automatic evaluation of summaries}. In \bibinfo{booktitle}{\emph{Text summarization branches out}}. \bibinfo{pages}{74--81}.
\newblock


\bibitem[Liu et~al\mbox{.}(2024)]%
        {liu2024deepseek}
\bibfield{author}{\bibinfo{person}{Aixin Liu}, \bibinfo{person}{Bei Feng}, \bibinfo{person}{Bing Xue}, \bibinfo{person}{Bingxuan Wang}, \bibinfo{person}{Bochao Wu}, \bibinfo{person}{Chengda Lu}, \bibinfo{person}{Chenggang Zhao}, \bibinfo{person}{Chengqi Deng}, \bibinfo{person}{Chenyu Zhang}, \bibinfo{person}{Chong Ruan}, {et~al\mbox{.}}} \bibinfo{year}{2024}\natexlab{}.
\newblock \showarticletitle{Deepseek-v3 technical report}.
\newblock \bibinfo{journal}{\emph{arXiv preprint arXiv:2412.19437}} (\bibinfo{year}{2024}).
\newblock


\bibitem[Maia et~al\mbox{.}(2018)]%
        {maia201818}
\bibfield{author}{\bibinfo{person}{Macedo Maia}, \bibinfo{person}{Siegfried Handschuh}, \bibinfo{person}{Andr{\'e} Freitas}, \bibinfo{person}{Brian Davis}, \bibinfo{person}{Ross McDermott}, \bibinfo{person}{Manel Zarrouk}, {and} \bibinfo{person}{Alexandra Balahur}.} \bibinfo{year}{2018}\natexlab{}.
\newblock \showarticletitle{Www'18 open challenge: financial opinion mining and question answering}. In \bibinfo{booktitle}{\emph{Companion proceedings of the the web conference 2018}}. \bibinfo{pages}{1941--1942}.
\newblock


\bibitem[Mei et~al\mbox{.}(2025)]%
        {mei2025survey}
\bibfield{author}{\bibinfo{person}{Lang Mei}, \bibinfo{person}{Siyu Mo}, \bibinfo{person}{Zhihan Yang}, {and} \bibinfo{person}{Chong Chen}.} \bibinfo{year}{2025}\natexlab{}.
\newblock \showarticletitle{A survey of multimodal retrieval-augmented generation}.
\newblock \bibinfo{journal}{\emph{arXiv preprint arXiv:2504.08748}} (\bibinfo{year}{2025}).
\newblock


\bibitem[Paperno et~al\mbox{.}(2016)]%
        {lambada}
\bibfield{author}{\bibinfo{person}{Denis Paperno}, \bibinfo{person}{Germ{\'{a}}n Kruszewski}, \bibinfo{person}{Angeliki Lazaridou}, \bibinfo{person}{Quan~Ngoc Pham}, \bibinfo{person}{Raffaella Bernardi}, \bibinfo{person}{Sandro Pezzelle}, \bibinfo{person}{Marco Baroni}, \bibinfo{person}{Gemma Boleda}, {and} \bibinfo{person}{Raquel Fern{\'{a}}ndez}.} \bibinfo{year}{2016}\natexlab{}.
\newblock \showarticletitle{The {LAMBADA} dataset: Word prediction requiring a broad discourse context}. In \bibinfo{booktitle}{\emph{Proceedings of the 54th Annual Meeting of the Association for Computational Linguistics, {ACL} 2016, August 7-12, 2016, Berlin, Germany, Volume 1: Long Papers}}. \bibinfo{publisher}{The Association for Computer Linguistics}.
\newblock
\urldef\tempurl%
\url{https://doi.org/10.18653/V1/P16-1144}
\showDOI{\tempurl}


\bibitem[Papineni et~al\mbox{.}(2002)]%
        {papineni2002bleu}
\bibfield{author}{\bibinfo{person}{Kishore Papineni}, \bibinfo{person}{Salim Roukos}, \bibinfo{person}{Todd Ward}, {and} \bibinfo{person}{Wei-Jing Zhu}.} \bibinfo{year}{2002}\natexlab{}.
\newblock \showarticletitle{Bleu: a method for automatic evaluation of machine translation}. In \bibinfo{booktitle}{\emph{Proceedings of the 40th annual meeting of the Association for Computational Linguistics}}. \bibinfo{pages}{311--318}.
\newblock


\bibitem[Sun et~al\mbox{.}(2024)]%
        {sun2024integrated}
\bibfield{author}{\bibinfo{person}{Yiding Sun}, \bibinfo{person}{Feng Wang}, \bibinfo{person}{Yutao Zhu}, \bibinfo{person}{Wayne~Xin Zhao}, {and} \bibinfo{person}{Jiaxin Mao}.} \bibinfo{year}{2024}\natexlab{}.
\newblock \showarticletitle{An integrated data processing framework for pretraining foundation models}. In \bibinfo{booktitle}{\emph{Proceedings of the 47th International ACM SIGIR Conference on Research and Development in Information Retrieval}}. \bibinfo{pages}{2713--2718}.
\newblock


\bibitem[Wang(2018)]%
        {wang2018glue}
\bibfield{author}{\bibinfo{person}{Alex Wang}.} \bibinfo{year}{2018}\natexlab{}.
\newblock \showarticletitle{Glue: A multi-task benchmark and analysis platform for natural language understanding}.
\newblock \bibinfo{journal}{\emph{arXiv preprint arXiv:1804.07461}} (\bibinfo{year}{2018}).
\newblock


\bibitem[White et~al\mbox{.}(2024)]%
        {livebench}
\bibfield{author}{\bibinfo{person}{Colin White}, \bibinfo{person}{Samuel Dooley}, \bibinfo{person}{Manley Roberts}, \bibinfo{person}{Arka Pal}, \bibinfo{person}{Ben Feuer}, \bibinfo{person}{Siddhartha Jain}, \bibinfo{person}{Ravid Shwartz-Ziv}, \bibinfo{person}{Neel Jain}, \bibinfo{person}{Khalid Saifullah}, \bibinfo{person}{Siddartha Naidu}, {et~al\mbox{.}}} \bibinfo{year}{2024}\natexlab{}.
\newblock \showarticletitle{LiveBench: A Challenging, Contamination-Free LLM Benchmark}.
\newblock \bibinfo{journal}{\emph{arXiv preprint arXiv:2406.19314}} (\bibinfo{year}{2024}).
\newblock


\bibitem[Zhu et~al\mbox{.}(2024)]%
        {zhu2024yulan}
\bibfield{author}{\bibinfo{person}{Yutao Zhu}, \bibinfo{person}{Kun Zhou}, \bibinfo{person}{Kelong Mao}, \bibinfo{person}{Wentong Chen}, \bibinfo{person}{Yiding Sun}, \bibinfo{person}{Zhipeng Chen}, \bibinfo{person}{Qian Cao}, \bibinfo{person}{Yihan Wu}, \bibinfo{person}{Yushuo Chen}, \bibinfo{person}{Feng Wang}, {et~al\mbox{.}}} \bibinfo{year}{2024}\natexlab{}.
\newblock \showarticletitle{Yulan: An open-source large language model}.
\newblock \bibinfo{journal}{\emph{arXiv preprint arXiv:2406.19853}} (\bibinfo{year}{2024}).
\newblock


\end{thebibliography}

\end{document}